\documentclass{article}

\usepackage{microtype}
\usepackage{graphicx}
\usepackage{subcaption}
\usepackage{booktabs} 

\usepackage[colorlinks=true, linkcolor=blue, citecolor=blue, urlcolor=blue]{hyperref}
\usepackage{url}
\usepackage{graphicx}
\usepackage{booktabs}
\usepackage{makecell}

\usepackage[preprint]{icml2026}

\usepackage{amsmath}
\usepackage{amssymb}
\usepackage{mathtools}
\usepackage{amsthm}

\usepackage[capitalize,noabbrev]{cleveref}

\theoremstyle{plain}

\theoremstyle{definition}

\theoremstyle{remark}

\usepackage[textsize=tiny]{todonotes}

\icmltitlerunning{When Should We Introduce Safety Interventions During Pretraining?}

\begin{document}

\twocolumn[
  \icmltitle{When Should We Introduce Safety Interventions During Pretraining?
}



  \icmlsetsymbol{equal}{*}

  \begin{icmlauthorlist}
    \icmlauthor{Dylan Sam}{yyy}
    \icmlauthor{Sachin Goyal}{yyy}
    \icmlauthor{Pratyush Maini}{yyy,comp}
    \icmlauthor{Alexander Robey}{yyy}
    \icmlauthor{J. Zico Kolter}{yyy}
  \end{icmlauthorlist}

  \icmlaffiliation{yyy}{Carnegie Mellon University}
  \icmlaffiliation{comp}{DatologyAI}

  \icmlcorrespondingauthor{Dylan Sam}{dylansam@andrew.cmu.edu}
  \icmlkeywords{Language Models, Safety, Pretraining}
  \vskip 0.3in
]



\printAffiliationsAndNotice{}  

\begin{abstract}

Prior work has shown that safety interventions applied during pretraining, such as removing and rephrasing harmful content, can substantially improve the robustness of the resulting models. 
In this paper, we study the fundamental question that prior work has overlooked: \textit{“When during pretraining should safety interventions be introduced?”}
We keep the underlying data sources and pretraining interventions fixed, varying the intervention start time (after 0\%, 20\%, or 60\% of pretraining tokens). 
We find that the optimal start time is not one-size-fits-all: with standard top-\emph{k} decoding, introducing interventions after a short initial phase of safe-only pretraining (20\%-60\%) often yields the strongest robustness, with the clearest benefits emerging \emph{after} downstream, benign finetuning. 
In contrast, for safety-aware inference, interventions starting from the beginning improve steerability towards safer generations. 
Finally, we observe that earlier interventions reshape internal representations: linear probes more cleanly separate safe vs harmful examples. 
Our results are the first to establish intervention timing as a key curriculum design choice for safety.

\end{abstract}

\section{Introduction}

Large language models (LLMs) are increasingly deployed in high-stakes settings, such as medical domains \citep{thirunavukarasu2023large, lievin2024can}, social sciences and simulations \citep{ziems2024can, byun2025using},  and in robotics \citep{firoozi2025foundation, robey2024jailbreaking}, where unsafe generations can cause real-world harms. Most production systems still rely on post-hoc alignment (e.g., RLHF~\citep{ouyang2022training}, DPO~\citep{rafailov2023direct}, or constitutional variants~\citep{bai2022constitutional}) that reshape surface model behavior after completing pretraining. 
Yet, a growing body of evidence suggests these methods are brittle: alignment effects are often restricted to the first few tokens of a response~\citep{qi2024safetyalignmentjusttokens} and are easily circumvented by benign finetuning \citep{qi2024safety} or adversarial pressure through prompting~\citep{zou2023universal, chao2023jailbreaking}. 
Misaligned behavior can be further elicited, and even amplified, by malicious or adversarial finetuning on a single task, which has been shown to generalize to other harmful tasks \citep{betley2025emergent}.

\begin{figure*}[t]
    \centering
    \includegraphics[width=0.7\linewidth]{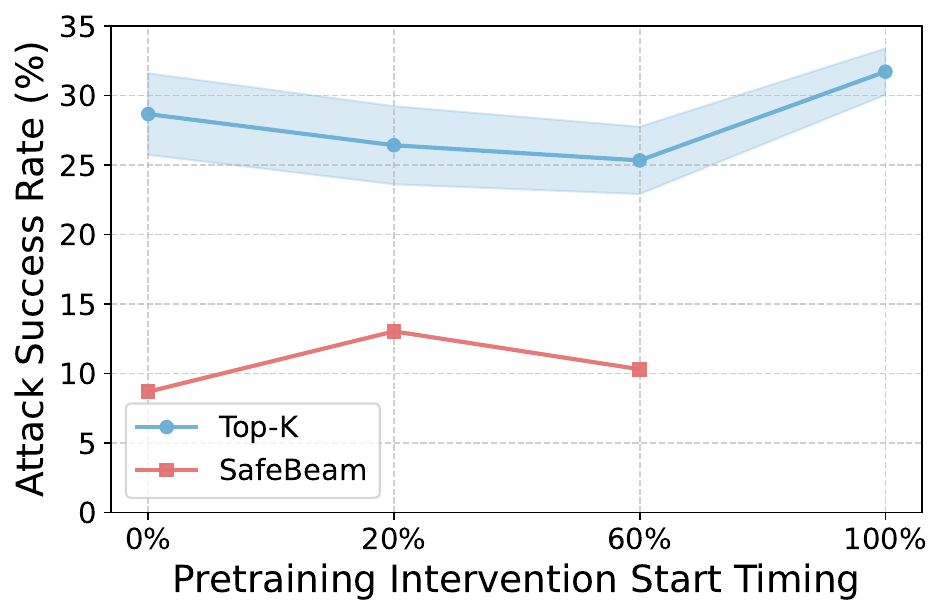}
    \caption{Attack success rate of the resulting \textbf{base models} as we vary the time at which we introduce safety pretraining interventions. For standard inference (Top-K), we report the shaded region as the standard deviation computed over 5 seeds. When using SafeBeam, the results are deterministic. Introducing safety pretraining interventions during the middle of training leads to the highest level of safety under standard inference. SafeBeam benefits from interventions being introduced at the start of pretraining.}
    \label{fig:base_results}
\end{figure*}

These vulnerabilities have motivated a shift toward integrating safety directly into pretraining.
Recent work has explored removing all harmful content from pretraining data via filtering approaches \citep{o2025deep}. 
Other work, however, has demonstrated that this could lead to a lack of understanding of harmful content, making models more prone to comply with harmful requests \citep{li2025when}. 
More promisingly, \citet{maini2025safety} explored synthetic pretraining interventions that maintain information from harmful examples while removing explicitly harmful content or providing sufficient context about why such content is harmful. 
These safety pretraining interventions show promise for improving refusal behavior, robustness to adversarial jailbreaks, and resilience to benign finetuning.

Despite this progress, all prior work on safety pretraining applies interventions uniformly from the start of training, implicitly assuming that \emph{when} safety signals are introduced does not matter. 
We challenge this assumption.
Drawing on insights from curriculum learning \citep{bengio2009curriculum, graves2017automated}, which has shown that the ordering of training examples can qualitatively change downstream performance, we ask: \textbf{when during pretraining should safety interventions be introduced?}
This question has natural intuitions on both sides: interventions introduced too late may be superficial and easily removed by downstream finetuning, while interventions introduced too early may degrade capabilities or over-regularize the model before it has learned basic linguistic structure.
To our knowledge, we are the first to systematically study this question.

In this paper, we empirically study the impacts of introducing safety interventions at different times during pretraining. We adopt the various pretraining interventions from the work of \citet{maini2025safety}, such as synthetic rephrasing and contextualization, metadata annotation, and refusal training, and introduce them at different stages (i.e., after 0\%, 20\%, or 60\% of the total amount of 600B tokens) of pretraining. This leads to various amounts of safety interventions token observed during pretraining.
Our experiments are run in the setting of pretraining a 1.7B parameter language model with the SmolLM2 architecture \citep{allal2025smollm2smolgoesbig}, on a pretraining mixture containing FineWeb-Edu \citep{penedo2024finewebdatasetsdecantingweb}, math, code, and other synthetic datasets \citep{allal2025smollm2smolgoesbig}. 
To evaluate how intervention timing affects safety, we analyze the resulting models at multiple stages (base models, instruction-tuned models, and models after benign finetuning).
We also study two common deployment choices: (a) standard top-\emph{k} decoding, and (b) safety-aware inference enabled by metadata. These settings prioritize different properties (robustness vs steerability), so the best intervention timing differs.

\begin{figure*}[t]
    \centering
    \includegraphics[width=0.7\linewidth]{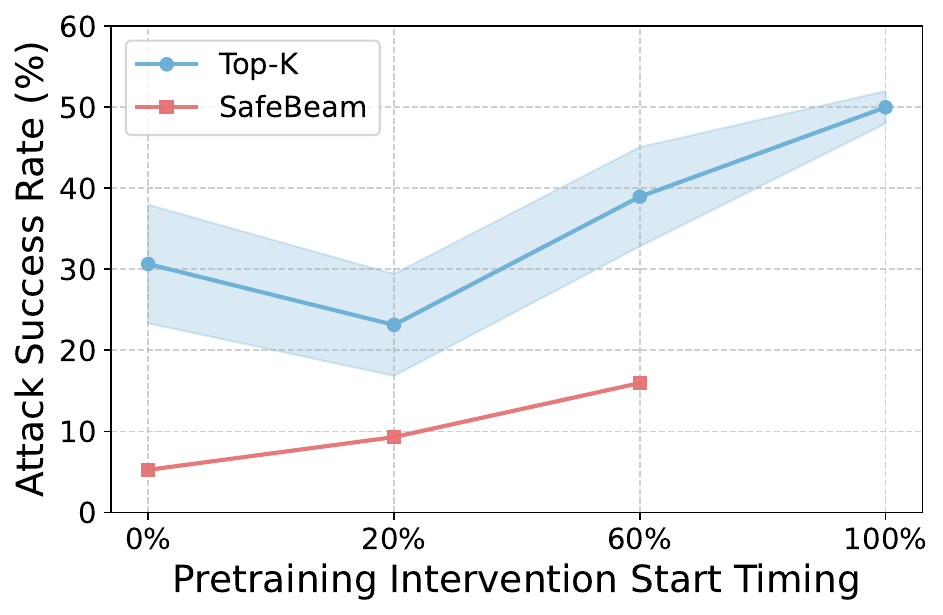}
    \caption{Attack success rate of the resulting \textbf{benign finetuned models on math data} (GSM8k) as we vary the time at which we introduce safety pretraining interventions. For standard inference (Top-K), we report the shaded region as the standard deviation computed over 5 seeds. When using SafeBeam, the results are deterministic. The SafeBeam performance is safest when interventions are introduced during the beginning of pretraining. Standard inference benefits from introducing safety interventions early (i.e., 0\% or 20\%) during pretraining.}
    \label{fig:gsm_results}
\end{figure*}

We find that the most effective start time depends on the evaluation setting and inference procedure. Under standard top-\emph{k} decoding, introducing interventions after some initial safe-only pretraining (after 20\%--60\%) often yields the safest base and instruction-tuned models, and provides the strongest robustness after benign finetuning.
In contrast, when using a safety-aware inference-time algorithm (SafeBeam), starting interventions from the beginning yields the most steerability toward safer generations.
This is complemented by our representation analysis: introducing interventions earlier makes safe vs harmful examples more linearly separable in the final-layer representation space, especially when metadata is included.
Finally, we find that intervening after an initial safe-only phase (20\% or 60\%) improves robustness to adversarial jailbreaks learned via GCG \citep{zou2023universal}, when compared to performing fully safe-only pretraining (100\%) and safety interventions applied from the start (0\%).
Overall, our results highlight curriculum timing as a key design choice: early starts improve steerability under safety-aware inference, while delayed starts can improve robustness under standard decoding and post-benign finetuning evaluation.

\section{Related Work}

\paragraph{Pretraining Safety}
Most language model pretraining employs simple heuristic filters that remove a small amount of harmful content, such as from particular URLs or explicit profanity \citep{raffel2020exploring, xue2020mt5}.
One early work studied the role of human preferences in language model pretraining~\cite{korbak2023pretraining}, finding that it yields lower toxicity scores relative to finetuned models (on the 100M parameter scale). 
Other works have taken a more explicit data curation approach. More recently, works have studied heuristic filtering based on educational content \citep{gunasekar2023textbooks, penedo2024finewebdatasetsdecantingweb}, which also correlates with (lower levels of) toxicity. 
Yet, \citet{vidgen2024introducing} note that harmful examples remain even in these educational datasets. 

As such, some work explores the impacts of filtering unsafe content from the pretraining corpus, training only on safe examples \citep{li2025when, o2025deep}.
Most recently and relatedly, the work of \citet{maini2025safety} studies the role of synthetic data generation to improve levels of safety in pretraining examples, while still retaining as much information as possible, finding large improvements in robustness after a round of benign finetuning. We expand upon this work to study the question of when such interventions should be deployed during pretraining.

\begin{figure*}[t]
    \centering
    \includegraphics[width=0.7\linewidth]{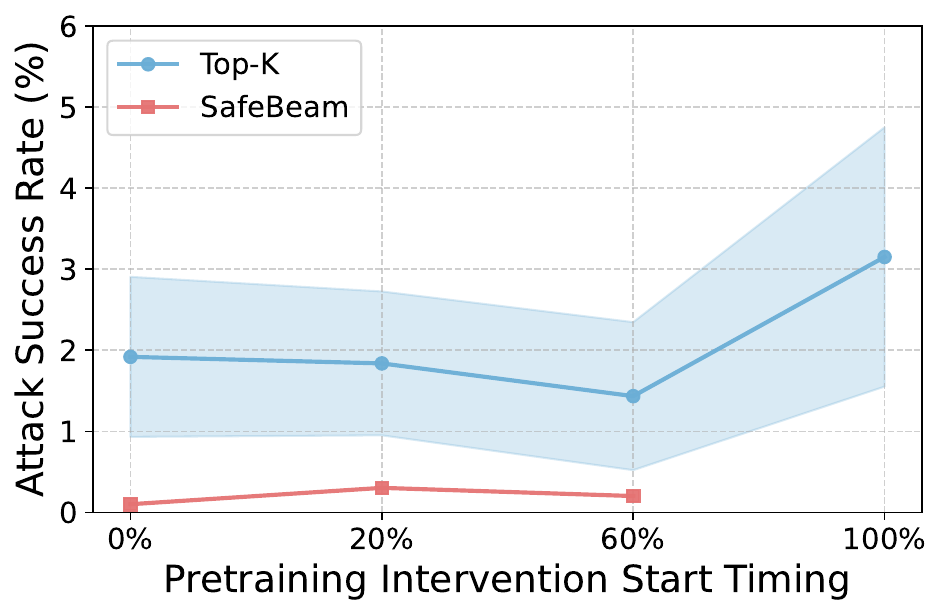}
    \caption{Attack success rate of the resulting \textbf{instruction-tuned models} as we vary the time at which we introduce safety pretraining interventions. For standard inference (Top-K), we report the shaded region as the standard deviation computed over 5 seeds. When using SafeBeam, the results are deterministic. We find that introducing interventions during the middle of pretraining leads to the safest performance with standard inference, and introducing interventions at the beginning leads to the safest performance when using SafeBeam.}
    \label{fig:ift_results}
\end{figure*}

\paragraph{Curriculum Learning}
Curriculum learning studies how the order and composition of training data shape generalization. Early work argues for presenting examples in a progression (e.g., from simpler to harder), and for adapting the sampling policy as training proceeds \citep{bengio2009curriculum, graves2017automated}. 
Such approaches have led to success in various settings, such as question-answering \citep{sachan2016easy} or reinforcement learning \citep{zhang2020automatic}.
Other work has explored when and what types of certain curriculum learning strategies work, demonstrated on small-scale image classification tasks \citep{wu2021when}.
Progress in the language model pretraining realm has been relatively static, likely due to the difficulty in measuring the hardness \citep{abdin2024phi} and similarity of content \citep{sam2025analyzing}, as well as the computational demands.

\paragraph{Stages of Language Model Pretraining} 
In large-scale language pretraining, most practical curricula have focused on adjusting mixture weights over data sources rather than token-level difficulty, largely due to computational complexity. For example, optimizing source mixtures during pretraining can yield gains in downstream performance and stability \citep{xie2023doremi}, and a line of work explores adaptive sampling or stage-specific mixtures that emphasize more useful distributions as training evolves \citep{jiang2024adaptive, ye2025data}. 

Complementary results in a more coarse-grained setting suggest that pretraining in distinct stages can change what models retain and how they transfer \citep{blakeney2024does}. 
Our study is aligned with this perspective but targets a safety-specific question: instead of asking which domains to upweight, we vary \emph{when} safety-augmented data enters pretraining. This allows us to measure whether safety signals introduced earlier versus later have a greater impact.

\section{Curricula for Safety}

This section formalizes our notion of a safety curriculum, detailing the particular interventions and how/when they are incorporated into the pretraining process.

\subsection{Safety Pretraining Interventions}

Crucial to this work are safety interventions that can be explored during the pretraining process, which are introduced in the work of \citet{maini2025safety}. 
We briefly introduce and describe these various pretraining interventions below.

\paragraph{Contextualized Rephrasing}
For pretraining examples that contain harmful content or instructions, the example is rewritten to preserve useful factual or domain knowledge while explicitly providing context that frames the material as risky or inappropriate to execute. 
The rewritten passages describe why the behavior is unsafe, when it is permissible to discuss at a high level, and how to redirect toward safer generations. 
The goal is to retain conceptual coverage (so the model “knows about” the topic) while discouraging unsafe execution, all within the same pretraining objective. 
Such examples are rephrased with a LLaMA3.1 8B parameter model \citep{touvron2023llama}.

\paragraph{Refusal Training}
The most harmful pretraining examples are converted with a similar synthetic data pipeline into conversation-style request and refusal pairs to incorporate throughout pretraining.
Prompts that elicit harmful actions are paired with refusals that model helpful-but-safe behavior: declining the request, briefly explaining safety rationales, and, when appropriate, offering a safe alternative or high-level guidance. 
These pairs are added directly to the pretraining corpus so that refusal behavior is learned natively to the model throughout pretraining, rather than only during a post-hoc alignment phase. This strengthens robustness towards downstream finetuning, where refusal behavior can easily be eroded either on benign or explicitly harmful tasks \citep{qi2024safety, betley2025emergent}.

\paragraph{Metadata-annotated Pretraining and SafeBeam}
Examples are flagged with a special token (e.g., ``\texttt{<potentially\_unsafe\_content>}''), which indicates that a pretraining example is harmful. As with the other interventions, this is implemented as ordinary text (randomly inserted with a 5\% chance at each of the token positions of the pretraining example) that the model predicts, avoiding changes to the training loss. 
These metadata annotations build associations within the model of harmful content and the special token, which enables steerability during inference-time. Specifically, this enables a SafeBeam search algorithm, which is a safety-focused inference-time modification. This modifies a standard beam search algorithm by dropping candidate beams that have a high probability of the next token being the harmful special token, leading to safer generations. The ultimate goal of such an intervention is to steer the model towards safer and high-quality completions \citep{yuan2025hard}.

We adopt the same resulting datasets from applying these interventions on the FineWeb dataset \citep{penedo2024finewebdatasetsdecantingweb} and other components of the SmolLM2 pretraining corpus \citep{allal2025smollm2smolgoesbig}.

\begin{table*}[t]
    \centering
    {\fontsize{10pt}{12pt}\selectfont  
    \setlength{\tabcolsep}{3pt}
    \renewcommand{\arraystretch}{1.15}
    \caption{\textbf{Capability evaluation results across intervention start timings.} We report accuracy on standard benchmarks: ARC-Challenge, CommonsenseQA (CS-QA), GSM8K, OpenBookQA, PIQA, TriviaQA, and Winogrande. We observe no obvious degradations in capabilities as we include safety pretraining interventions earlier (exposing models to more safety pretraining data); in fact, earlier pretraining interventions seem to lead to higher levels of capabilities.}
    \label{tab:capability_evals}

    \begin{tabular}{l|ccccccccc}
    \toprule
    \textbf{Intervention} &
    \textbf{Avg.} & \textbf{ARC-C} & \textbf{CS-QA} & \textbf{GSM8K} &
    \textbf{OpenBookQA} &
    \textbf{PIQA} & \textbf{TriviaQA} & \textbf{Wino.} \\
    \midrule
    0\%   & 37.2\% & 42.9\% & 19.7\% & 7.4\% & 42.6\% & 76.6\% & 12.7\% & 58.8\%  \\
    20\%  & 36.7\% & 43.5\% & 22.9\% & 6.1\% & 39.8\% & 76.0\% & 10.0\% & 58.7\%  \\
    60\%  & 35.6\% & 41.8\% & 21.1\% & 5.5\% & 41.6\% & 75.1\% & 5.8\%  & 58.0\%  \\
    100\% & 36.1\% & 42.9\% & 21.6\% & 4.9\% & 42.8\% & 75.1\% & 6.0\%  & 59.4\% \\
    \bottomrule
    \end{tabular}
    } 
\end{table*}

\subsection{Safety Curricula}

A central question of this work is how the timing of safety interventions during pretraining shapes downstream robustness. Prior work on safety pretraining \citep{maini2025safety} has introduced interventions from the very start of training---effectively modifying the full corpus before pretraining begins. 
Other work has demonstrated that orderings of certain concepts can improve the speed of learning \citep{chen2023skill}.
Similarly, in the human learning process, we often first learn about general knowledge, such as language and arithmetic, before addressing potentially harmful or controversial content, such as wars and tragedies. At the other extreme, filtering approaches \citep{o2025deep} that permanently and completely remove all harmful content during pretraining potentially lose too many capabilities from the underlying model. We adopt a similar approach here, where we study a curriculum of first pretraining for a phase on safe-only data, before introducing safety interventions after a certain percentage of tokens of the full 600B token pretraining budget.

To study the role of the \textit{timing} of introducing safety interventions, we study 4 main settings. 
The first setting (0\%) applies interventions from the very beginning, which directly mirrors the setup in prior Safety Pretraining work \citep{maini2025safety}. This captures no pretraining phase on safe-only data. 
The second setting (20\%) delays interventions until after the model has already seen a substantial portion of ``safe'' general web-scale data, allowing it to acquire broad linguistic and factual knowledge before any safety interventions. 
The third setting (60\%) introduces interventions only in the later stages of training, when much of the model’s base distribution of language has already been established. 
Finally, the last setting (100\%) corresponds to fully training on safe-only data, which is equivalent to prior work that filters out all unsafe content \citep{o2025deep}.

To also vary the types of interventions employed, we explore the two primary settings from Safety Pretraining \citep{maini2025safety}: (i) we first study using all of the interventions in conjunction, including the metadata tagging and inference-time modifications, (ii) using only rephrasing and refusal training without any metadata tagging. The second set of experiments here considers a slightly less aggressive modification to pretraining, only in terms of data curation.

\begin{figure*}[t]
    \centering
    \includegraphics[width=0.75\linewidth]{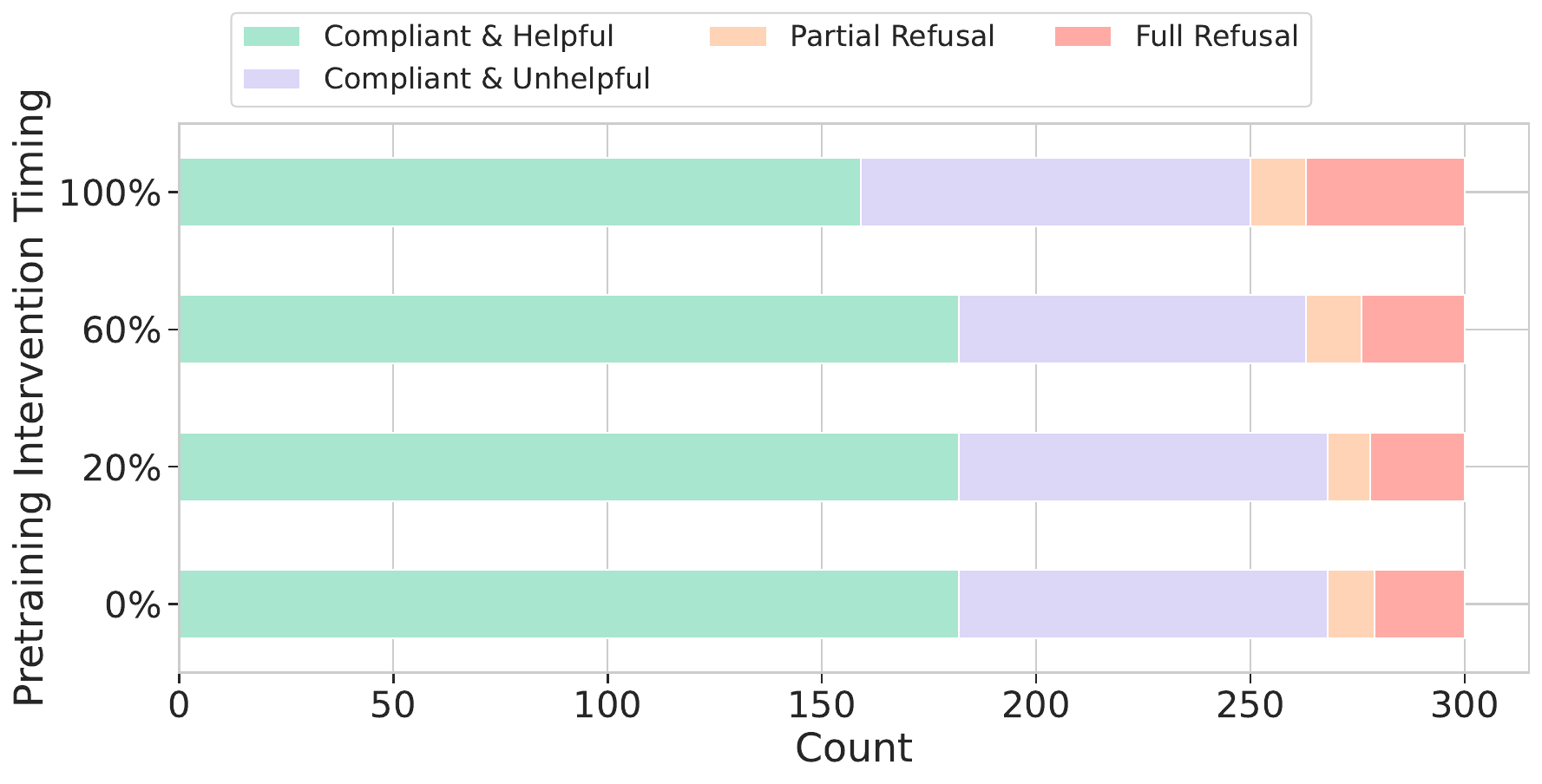}
    \caption{Comparison of overrefusal rates on Alpaca \citep{taori2023stanford} when using standard (Top-k) sampling for inference. We find that \textbf{incorporating interventions earlier during pretraining leads to a better compliance rate on benign requests}. }
    \label{fig:helpfulness_topk}
\end{figure*}

\section{Experiments}

To evaluate the impacts of the timing of safety curricula, we adopt evaluations along multiple axes of the model pipeline. Our evaluations include safety (of base model, instruction-tuned models, and finetuned models), helpfulness on harmless and benign queries, jailbreakability through adversarial suffixes, and in analyzing model internals and the underlying representation space.

\subsection{Safety Evaluations}
We follow the work of \citet{maini2025safety}, performing safety evaluations at 3 main stages: (i) base model -- immediately after pretraining, (ii) instruction-tuned models -- after a standardized post-training setup, and (iii) benign finetuned models -- after a round of benign finetuning on math data, e.g., GSM8K \citep{cobbe2021training}.

We first evaluate the safety of our models immediately after pretraining. We use the completion-style harmful request benchmark from \citet{maini2025safety}. This evaluates the tendencies of the base models to complete the beginning of a harmful prompt. 
Next, we evaluate the models after performing a standard instruction tuning recipe, which also includes some safety training and refusal examples. 
Our evaluation is comprised of 4 standard safety benchmarks: \texttt{HarmBench} \citep{mazeika2024harmbench}, \texttt{TDC} \citep{maloyan2024trojan}, \texttt{JailbreakBench} \citep{chao2024jailbreakbench}, \texttt{AdvBench} \citep{zou2023universal}. 
We also incorporate a harder evaluation setting that incorporates adversarial jailbreaking, where we learn adversarial suffixes via the GCG algorithm \citep{zou2023universal} that are optimized on each model to increase the rate of compliance with harmful requests.

Finally, we evaluate the models after an epoch of benign, supervised finetuning on a helpful, non-adversarial dataset of GSM8K \citep{cobbe2021training}, which is common practice for mathematical reasoning, without any focus on safety or alignment. 
Despite this being non-adversarial data, other works have demonstrated that such rounds of supervised finetuning can erode aligned behaviors \citep{qi2024safety, betley2025emergent}. 

\subsection{Capability and Helpfulness Evaluations}

When introducing safety interventions that may increase refusal rates, it is crucial to assess any impact on model capabilities and helpfulness. For capabilities, we evaluate on a suite of standard benchmarks spanning general knowledge (ARC-Challenge, TriviaQA), commonsense reasoning (CommonsenseQA, OpenBookQA, PIQA, Winogrande), and mathematical reasoning (GSM8K). Additional experiment details are deferred to Appendix \ref{appx:exp_details}.

To measure helpfulness, we evaluate model compliance on benign requests from the Alpaca dataset~\citep{taori2023stanford}. We use GPT-4o-mini as a judge to classify each response into one of four categories: (1) compliant and helpful, (2) compliant but unhelpful, (3) partial refusal, or (4) full refusal. Categories (1) and (2) represent desirable compliance behavior, while categories (3) and (4) indicate overrefusal. For models using the SafeBeam algorithm, we additionally introduce a fifth category: (5) shortcircuit, which occurs when generation terminates upon producing the harmful metadata token. In practice, such responses can be post-processed at serving time by substituting a default refusal message from a set of known safe completions.

\begin{figure*}[t]
    \centering
    \includegraphics[width=0.7\linewidth]{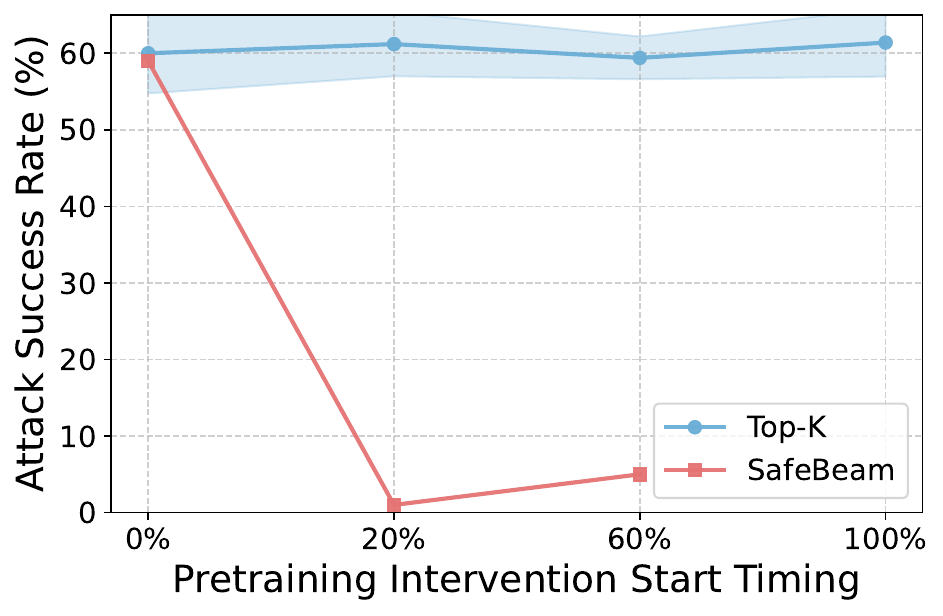}
    \caption{Jailbreakability of our models via the GCG algorithm \citep{zou2023universal} as we vary the timing of safety pretraining interventions. We find that introducing interventions during the middle stages of pretraining greatly improves the steerability of models towards safe completions and refusals when exposed to adversarially learned suffixes.}
    \label{fig:gcg_timing}
\end{figure*}

\subsection{Representation Analysis}
To analyze the impact of various curricula on the underlying representation space of our models, we evaluate the separability of safe vs harmful content. 
We train linear probes to classify between examples from FineWeb data that have been annotated as safe versus unsafe, learned over the representations taken from the internal activations of the model's final layer. For the safety label, we use the safety score from \citet{maini2025safety}, which is produced by GPT-4o-mini that judges pretraining examples with a safety rubric.

\subsection{Training Details}

\paragraph{Pretraining Setup~} 
For our experiments, we adopt the same setup of the SmolLM2~\citep{allal2025smollm2smolgoesbig} for all our experiments. We train 1.7B parameter language models with a pretraining corpus of a mixture of \texttt{FineWeb-Edu}, \texttt{StackOverflow}, \texttt{FineMath}, and \texttt{Cosmopedia}. 
For our safety pretraining interventions, we adopt the modified datasets of \texttt{SafeWeb}, \texttt{RefuseWeb}, and \texttt{MoralEducation} from \citet{maini2025safety}, which are safety-focused, rephrased and recontextualized versions of harmful pretraining examples from \texttt{FineWeb-Edu}.

Pretraining is performed using \texttt{LitGPT}~\citep{litgpt-2023}, with FlashAttention-2 enabled and mixed-precision training for efficiency.
We use the same hyperparameters (e.g., learning rate schedule, batch size, and sequence length) as in the SmolLM2 pretraining setup.

\paragraph{Post-training Setup~}
For a standardized post-training stack, we use a combination of \texttt{Ultrachat-200k} with \texttt{AllenAI WildGuardMix} and \texttt{WildJailbreak} datasets for safety instruction tuning. 
This follows standard practice from prior works in safety training~\citep{zou2024improvingalignmentrobustnesscircuit, maini2025safety}.

\section{Results}

We now present our findings on how safety curricula influence robustness, helpfulness, and internal representations.

\begin{figure*}[t]
    \centering
    \includegraphics[width=0.49\linewidth]{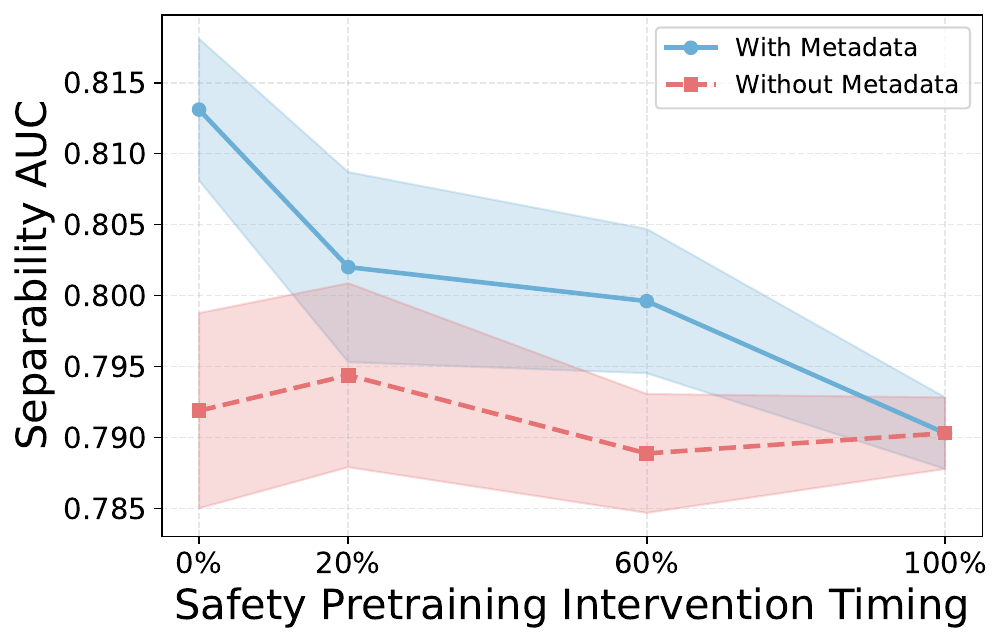}
    \includegraphics[width=0.49\linewidth]{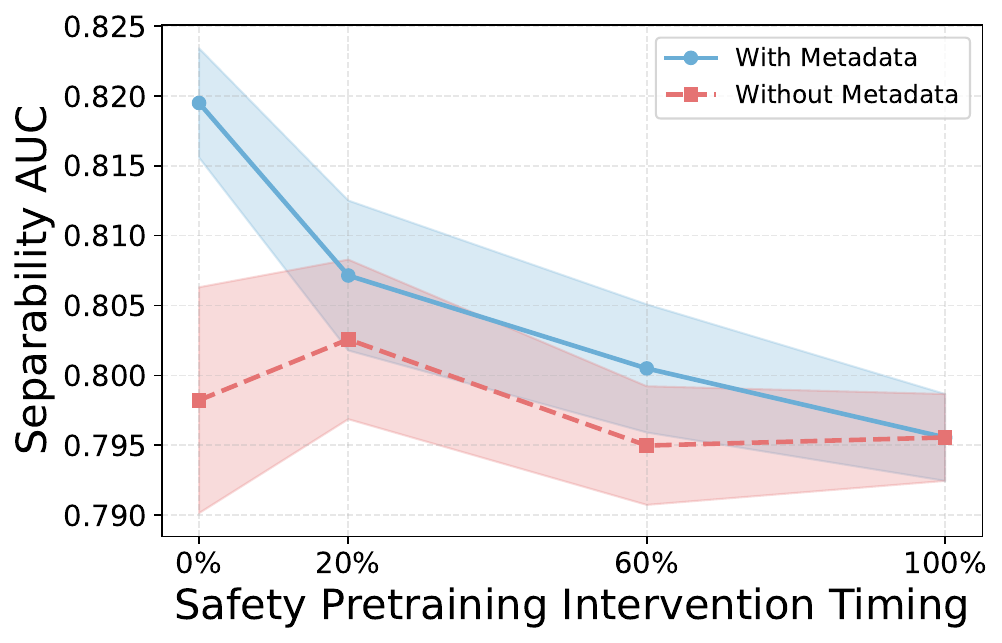}
    \caption{Separability of safe vs unsafe content (as annotated by GPT-4o-mini) in the \textbf{last layer representation space} of base models (Left) and instruction-tuned models (Right), as we vary the start timing when we incorporate safety pretraining interventions. We find that \textbf{introducing interventions earlier yields greater separability in representation space} (especially with the incorporation of harmful metadata tags). The shaded regions denote the standard deviation computed over 5 training runs.}
    \label{fig:separability}
    \vspace{-2mm}
\end{figure*}

\subsection{Safety Results}

\paragraph{Base Models} Timing matters even before instruction tuning (Figure~\ref{fig:base_results}). Under standard top-k sampling, introducing interventions in the middle of pretraining (20\% or 60\%) yields the safest base models, while training on safe-only data (100\%) is the least safe. This again complements the findings from prior work \citep{li2025when, maini2025safety}.
In contrast, with SafeBeam, incorporating interventions and metadata from the start (0\%) provides measurable gains.

\paragraph{Instruction-tuned Models} A similar pattern holds with the models that have undergone instruction tuning (Figure~\ref{fig:ift_results}).  
Models trained with interventions during the middle of pretraining exhibit stronger safety performance under top-\emph{k} inference than those with no interventions. 
Again, SafeBeam benefits the most when metadata is incorporated from the beginning (i.e., 0\%).

\paragraph{Benign Finetuned Models} 
The clearest and most noticeable effects emerge in our evaluation setting after a benign finetune on mathematical data (Figure~\ref{fig:gsm_results}). 
Here, SafeBeam achieves the greatest robustness when interventions are included from the very start of pretraining. For standard top-\emph{k} decoding, introducing interventions after roughly 20\% of training yields the strongest post-finetuning safety.  

Taken together, our evaluations suggest two main takeaways. First, when using modified SafeBeam inference, introducing metadata-annotated unsafe data earliest in pretraining leads to the best steerability towards safer completions. Secondly, with top-k inference, introducing safety pretraining interventions during the middle of training (after 20\% or 60\%) leads to the most robust performance. This suggests that allowing the model to first absorb general-purpose safe data before applying any safety interventions may produce alignment that is more robust to subsequent benign finetuning.

\subsection{Capability Results}

We evaluate whether safety pretraining interventions degrade general model capabilities across various standard benchmarks (Table~\ref{tab:capability_evals}). 
Overall, we find that safety interventions introduce minimal capability trade-offs and, in some cases, may even improve performance. 
We attribute this to the model's efficient learning from high-quality synthetic data~\citep{maini-etal-2024-rephrasing}. 
Taken together, these results indicate that safety pretraining interventions do not meaningfully degrade model capabilities; indeed, earlier interventions appear mildly beneficial on average.

\subsection{Overrefusal Results}

We assess the impact of pretraining interventions on model helpfulness using benign requests from the Alpaca benchmark~\citep{taori2023stanford}. 
Under standard top-$k$ sampling (Figure~\ref{fig:helpfulness_topk}), incorporating interventions earlier during pretraining (0\% or 20\%) yields a slight improvement in compliance rates. These results suggest that safety pretraining interventions introduce minimal trade-offs in overrefusal behavior under standard inference.

When using the SafeBeam algorithm (Figure~\ref{fig:overrefusal_safebeam}), we observe a more nuanced pattern: earlier interventions produce modestly higher overrefusal rates, yet the quality of compliant responses improves. The figure reveals that earlier pretraining interventions (0\% and 20\%) increase the fraction of compliant-and-helpful responses among non-refused outputs, even as overall refusal rates rise slightly. We hypothesize that this reflects improved calibration: early interventions strengthen the correspondence between the harmful metadata tag and genuinely unsafe content, enabling SafeBeam to filter more precisely. The surviving beams thus tend to produce higher-quality completions. 



\subsection{Jailbreaking Results}

We also present results on how susceptible our models are to jailbreaking attacks using the GCG algorithm \citep{zou2023universal}, as we vary the timing of the safety pretraining interventions. Surprisingly, we find that introducing pretraining interventions after some initial safe-only pretraining (i.e., at 20\% or at 60\%) leads to \textbf{significantly more robustness} to jailbreaks via GCG when using the SafeBeam inference algorithm (Figure \ref{fig:gcg_timing}). 
On the other hand, we find that jailbreakability under standard inference does not noticeably change given different intervention timings.

\subsection{Representation Results}

In settings without harmful metadata tags, earlier interventions slightly improve the separability of safe versus unsafe points (with a linear classifier over the last-layer activations).  
When metadata is included, the effect is more pronounced: introducing interventions from the start yields the highest linear-probe AUC, suggesting a much cleaner separation of harmful and safe content in the representation space. This also supports the improved SafeBeam robustness, where models are more steerable towards safer responses when interventions are included early in pretraining.

\section{Discussion}

Our work studies the role of different safety curricula, or how the timing of safety pretraining interventions (e.g., when and how safety-augmented data enters pretraining) impacts the robustness of our models. 
Our experiments show that incorporating safety pretraining interventions indeed help, but the best timing depends on the setting: delayed interventions (e.g., 20\%--60\%) can yield stronger robustness under standard decoding and against adversarial jailbreaks, while early interventions (0\%) improve steerability under safety-aware inference (SafeBeam) and yield clearer separation of safe vs unsafe content in representation space. Taken together, these findings highlight how the timing of safety interventions is an important choice in shaping a model’s behavior, offering guidance for building more trustworthy and responsible language models.

\paragraph{Why does timing matter?} While our experiments establish that intervention timing significantly affects outcomes, understanding the underlying mechanisms remains an open question. We offer several hypotheses for future investigation.
For standard top-k decoding, delayed interventions (20\%-60\%) may succeed because the model benefits from first acquiring general linguistic competence and world knowledge before safety signals are introduced. Early safety interventions might compete with or interfere with learning basic representations, potentially causing the model to develop shallow safety heuristics rather than robust understanding. By contrast, when the model has already formed stable representations of language and concepts, safety interventions may integrate more deeply into these existing structures, making them harder to remove through subsequent finetuning.

For SafeBeam inference, early interventions may help because the metadata-based steering mechanism relies on strong, well-calibrated associations between the harmful content tag and genuinely unsafe material. Introducing these associations from the start of training allows them to be reinforced throughout the entire pretraining process, resulting in more reliable probability estimates that SafeBeam can exploit. This interpretation is supported by our representation analysis (Figure 6), which shows that early interventions with metadata yield the cleanest separation between safe and unsafe content in representation space.

A limitation of our study is that the introduction of safety pretraining interventions at different times during pretraining means that certain models have seen more synthetic contextualized data or refusal data. 
However, we note our main effects are \textbf{not monotonic} in the number of intervention tokens (i.e., “more intervention tokens” do not consistently yield safer models), suggesting that \emph{when} safety-augmented data is introduced matters beyond total quantity.
\clearpage




\bibliography{main}
\bibliographystyle{plainnat}
\clearpage

\appendix

\onecolumn

\section{Additional Experiments}
\label{sec:additional_experiments}

\begin{figure}[t]
    \centering
    \includegraphics[width=0.49\linewidth]{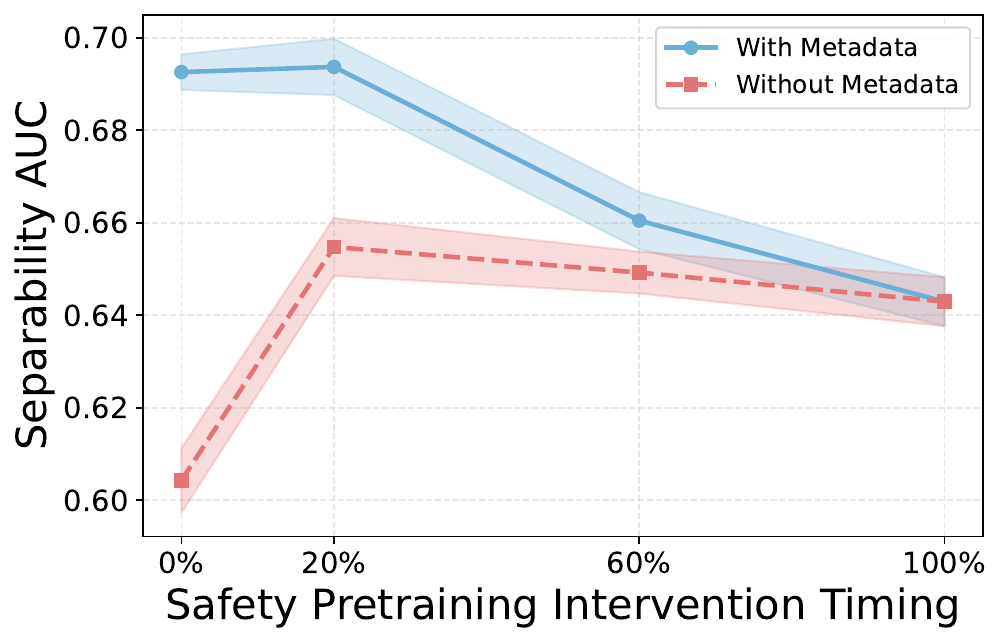}
    \includegraphics[width=0.49\linewidth]{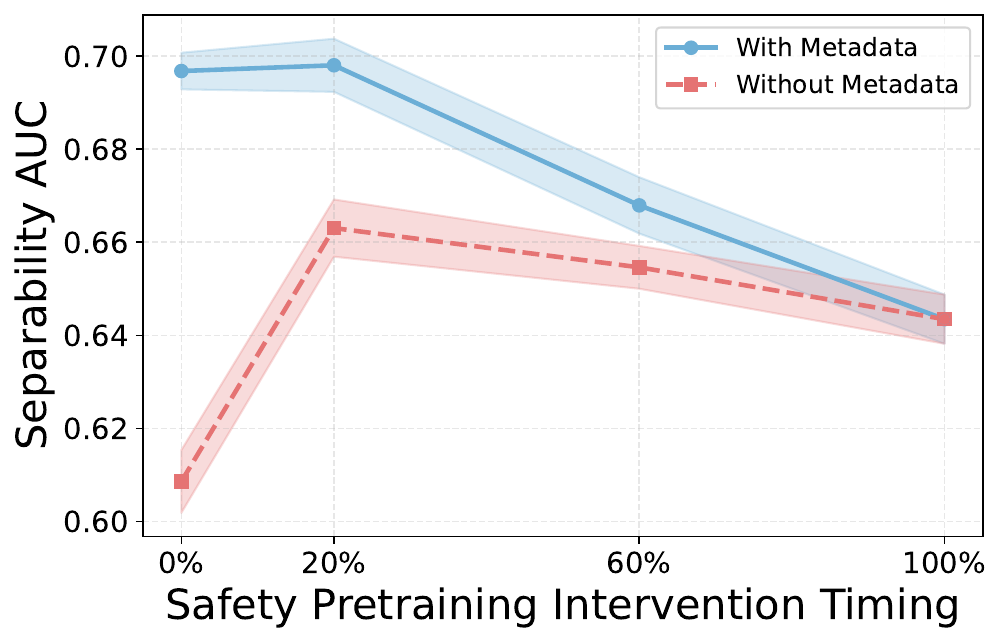}
    \caption{Separability of safe vs unsafe content (as annotated by GPT-4o-mini) in the \textbf{first} layer of base models (Left) and instruction-tuned models (Right), as we vary the start timing when we incorporate safety pretraining interventions. The shaded regions denote the standard deviation computed over 5 training runs.}
    \label{fig:separability_first}
\end{figure}

\begin{figure}[t]
    \centering
    \includegraphics[width=0.49\linewidth]{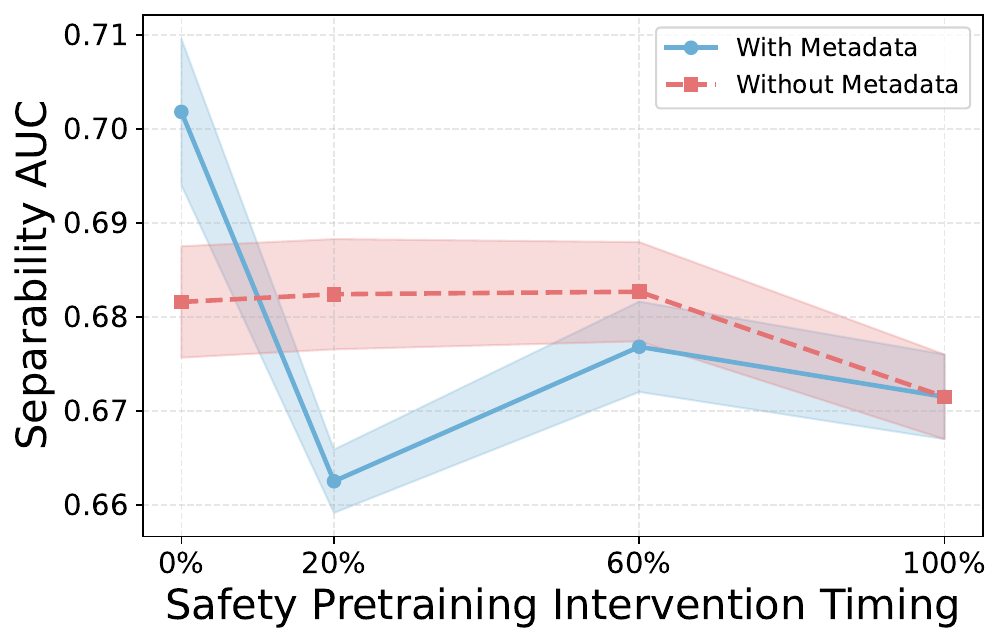}
    \includegraphics[width=0.49\linewidth]{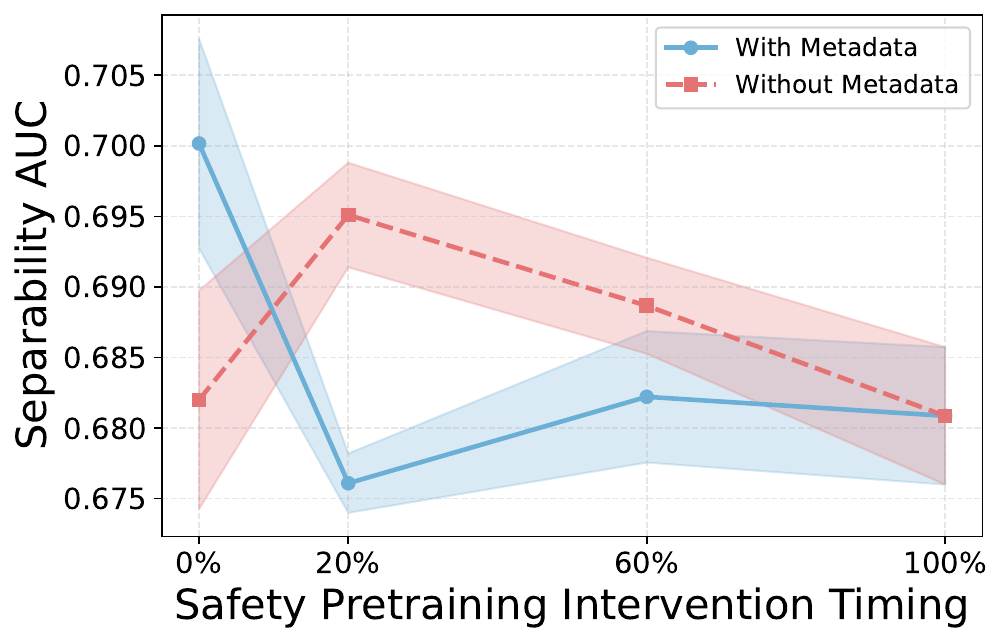}
    \caption{Separability of safe vs unsafe content (as annotated by GPT-4o-mini) in the \textbf{middle (13th) layer} of base models (Left) and instruction-tuned models (Right), as we vary the start timing when we incorporate safety pretraining interventions. The shaded regions denote the standard deviation computed over 5 training runs.}
    \label{fig:separability_middle}
\end{figure}

\begin{figure*}[t]
    \centering
    \includegraphics[width=0.75\linewidth]{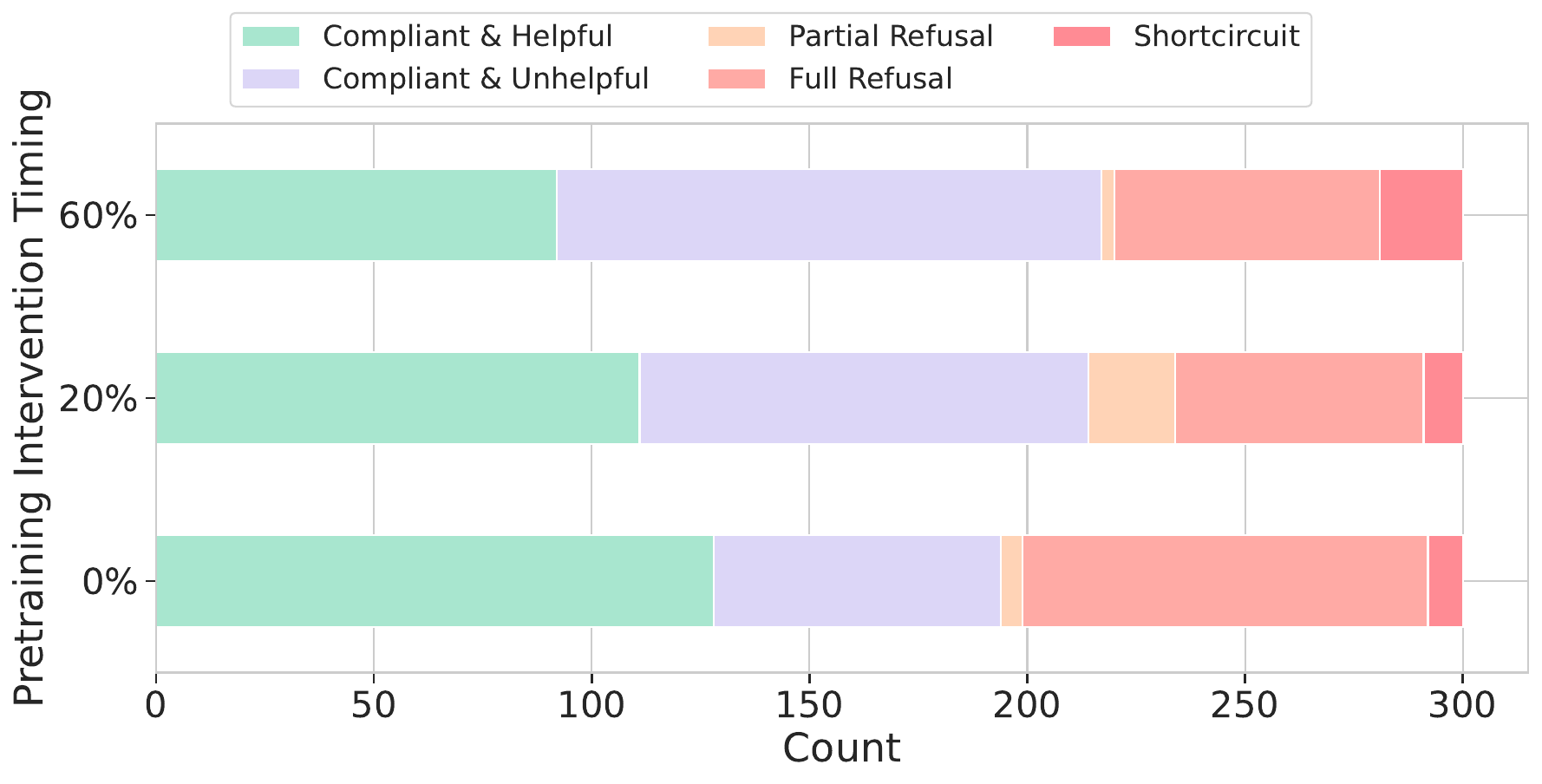}
    \caption{Comparison of overrefusal rates on Alpaca when using the SafeBeam inference algorithm. We find that \textbf{introducing pretraining interventions and metadata earlier during pretraining leads to more helpful compliant behavior, with a slight increase in overrefusal rates}.}
    \label{fig:overrefusal_safebeam}
\end{figure*}

\paragraph{Additional Separability Results.} In the main text, we present results on separability of harmful vs safe content in the final layer of the resulting pretraining models. Here, we also present results for the first layer and the middle (13th) layer. We observe similar trends, although we find greater separability in the final layer when compared to earlier layers in the models.

\paragraph{SafeBeam Overrefusal.}

Figure~\ref{fig:overrefusal_safebeam} extends our overrefusal analysis to the SafeBeam inference setting. Compared to standard top-$k$ sampling, SafeBeam introduces a fifth response category (\emph{shortcircuit}), which occurs when generation terminates upon producing the harmful metadata token. In deployment, such responses can be handled by substituting a predetermined safe completion.

\section{Additional Experimental Details}\label{appx:exp_details}

\paragraph{Safety Evaluation} For our standard inference evaluations, we use top-k sampling with $k = 10$ and truncate after 256 tokens. We average the results over 5 samples. For SafeBeam evaluation, we run a beam search with $10$ beams for 256 tokens, and discard the top 50\% of unsafe beams (in terms of probability of the next token being the harmful metadata token). We follow the same exact setting as \citet{maini2025safety}.
We adopt the same GPT-4o-mini judge and judging prompt for helpfulness evaluations, and the same HarmBench judge and judge prompt \citep{mazeika2024harmbench} for our safety evaluations.

\paragraph{Capabilities Evaluation} 
We use the default experimental setting for all the datasets as provided as in the LM evaluation harness \citep{eval-harness}. We use the dataset default few-shot settings: \textbf{0-shot} for ARC-Challenge, OpenBookQA, CommonsenseQA, PIQA, TriviaQA, and WinoGrande (\texttt{num\_fewshot}=0), and \textbf{5-shot} for GSM8K (\texttt{num\_fewshot}=5; few-shot examples sampled from the training split).

\paragraph{Adversarial Jailbreaking} For computing the adversarial suffixes via the GCG algorithm, we learn an adversarial suffix on the target model, optimizing for the target string of ``\texttt{Sure}''. We perform optimization for 500 steps, with an initialization of ``\texttt{x x ... x}'', and \texttt{top\_k} of 256.

\paragraph{Finetuning} 
For our supervised instruction tuning, we finetune with a batch size of 4 with a learning rate of 2e-5 for one epoch, using cosine learning rate decay with a warmup ratio of 0.03.
For our benign finetuning experiments, we perform supervised finetuning on GSM8k for a single epoch with a learning rate of 2e-5 with the same learning rate schedule as in instruction tuning.

\paragraph{Separability Experiments}
In our separability experiments, we use the last token position and the final layer of all models. For each model, we train a logistic regression classifier on a balanced subset of 500 safe and 500 unsafe examples from the training split, and the trained probe is evaluated on the full balanced test set. We report the mean, standard deviation, and standard error, when computed over five seeds, of the test AUC (area under the ROC curve).

\section{Additional Discussion}
We note that the notion of safety that we focus on in this paper is fairly broad, encompassing general notions of toxicity and harms. This is taken from the taxonomy proposed by the MLCommons \citep{vidgen2024introducing}, captures a wide variety of potential harms and undesirable behaviors in different domains. 
Our method likely requires additional efforts to tackle safety for specific capabilities, such as bio-weapons and cybersecurity \citep{li2024wmdp} or reliability of agents in context-dependent scenarios \citep{andriushchenko2025agentharm, sam2025evaluating}.

We also note that our results on the separability of safe vs unsafe content has greater implications for the monitoring of such models. These experiments contribute to a line of work that develops external monitoring to predict when undesirable generations will occur and steer towards safer outputs \citep{turner2023steering, zou2024improvingalignmentrobustnesscircuit}. 
This may even have implications for complementary alternative approaches that are adopted in black-box settings and use external models \citep{sharma2025constitutional, sam2025predicting}, which is an area for future study. 

\section{Computational Resources}

For each variant of the pretraining runs for the 1.7B parameter model on 600B tokens, we used 32 H100 GPUs for roughly 6-7 days. 
Finetuning requires roughly 4 H100 GPUs for 4 hours. 
Evaluations are run on one L40S GPU.

\end{document}